\documentclass[article]{jdssv}
\usepackage{graphicx} 
\usepackage{amsmath,amsfonts,mathtools,amssymb}
\usepackage{booktabs}
\usepackage{adjustbox}
\usepackage{algorithm} 
\usepackage{algpseudocode}

\usepackage{anyfontsize} 
\usepackage{silence} 
\DeclareFontFamilySubstitution{TS1}{aer}{cmr}
    
\definecolor{blue}{RGB}{0,0,255}
\definecolor{red}{RGB}{255,0,0}

\newcommand{\med}{\mathop{\mbox{med}}}
\newcommand{\MAD}{\mathop{\mbox{MAD}}}
\newcommand{\B}{\mathop{\mathcal{B}}}
    
\newcommand{\bef}{\boldsymbol{f}}
\newcommand{\bhf}{\boldsymbol{\hat{f}}}
\newcommand{\bhfi}{\boldsymbol{\hat{f}}\!_i}
\newcommand{\bhfj}{\boldsymbol{\hat{f}}\!_j}
\newcommand{\bx}{\boldsymbol{x}}
\newcommand{\by}{\boldsymbol{y}}
\newcommand{\bv}{\boldsymbol{v}}
\newcommand{\R}{\mathbb{R}}
\newcommand{\bF}{\mathbf{F}}
\newcommand{\bK}{\mathbf{K}}
\newcommand{\bL}{\boldsymbol{\Lambda}}
\newcommand{\bT}{\mathbf{T}}
\newcommand{\bU}{\mathbf{U}}
\newcommand{\bV}{\mathbf{V}}
\newcommand{\bX}{\mathbf{X}}
\newcommand{\SDO}{\mbox{SDO}}
    
\author{Can Hakan Da\u{g}{\i}d{\i}r\\KU Leuven
       \And Mia Hubert\\KU Leuven
       \And Peter J. Rousseeuw\\KU Leuven\\}
\Plainauthor{Can Hakan Dağıdır, Mia Hubert, 
             Peter J. Rousseeuw}
    
\title{Kernel Outlier Detection}
\Plaintitle{Kernel Outlier Detection}
\Shorttitle{Kernel Outlier Detection}
    
\Abstract{A new anomaly detection method 
called kernel outlier detection (KOD) is proposed. 
It is designed to 
address challenges of outlier detection in 
high-dimensional settings. The aim is to 
overcome limitations of 
existing methods, such as dependence on distributional 
assumptions or on hyperparameters that are hard to tune.
KOD starts with a kernel transformation, followed by
a projection pursuit approach. Its novelties include a new
ensemble of directions to search over, and a new way to
combine results of different direction types.
This provides a flexible and lightweight approach for 
outlier detection. Our empirical evaluations illustrate 
the effectiveness of KOD on three small datasets with 
challenging structures, and on four large benchmark 
datasets.}
    
\Plainkeywords{Anomaly detection, Outlyingness, Kernel 
transformation, Projection depth}
    
\Address{
Can Hakan Da\u{g}{\i}d{\i}r, Mia Hubert, Peter J. Rousseeuw\\
Section of Statistics and Data Science\\
Department of Mathematics, KU Leuven\\
Celestijnenlaan 200B\\
BE-3001 Leuven, Belgium\\
E-mail: \email{canhakan.dagidir@kuleuven.be}\\
E-mail: \email{mia.hubert@kuleuven.be}\\
E-mail: \email{peter.rousseeuw@kuleuven.be}\\
URL: \url{https://wis.kuleuven.be/statdatascience/robust}
}
    
\begin{document}
    
\maketitle
    
\section{Introduction} \label{sec:introduction}
    
Most outlier detection methods for multivariate and 
possibly high-dimensional data assume that the non-outlying 
data are drawn from an elliptical distribution. However, 
many modern datasets such as images, sensor data or genetic 
data do not obey this assumption. In this work, we propose 
a kernel outlier detection method that does not rely on this 
assumption, by combining kernel transformations with the 
Stahel-Donoho outlyingness 
\citep{stahel1981robust, donoho1982breakdown}. 
    
The Stahel-Donoho outlyingness (SDO) assigns an outlyingness 
value to each data point by considering the most extreme 
standardized deviation across all possible projection 
directions. In other words, it searches for the direction 
along which the observation deviates the most from the 
majority of the data. Formally, given a data matrix 
$(\bx_1, \dots, \bx_n)^T$ with $p$-variate data points 
$\bx_i = (x_{i1}, \dots, x_{ip})^T$, the SDO of each case 
is defined as
\begin{equation} \label{eq:sdo}
   \text{SDO}(\bx_i) \coloneq \sup_{\boldsymbol{v} \in \B} 
   \frac{\left| \bv^T \bx_i  - \med_j(\bv^T\bx_j) \right|}
   {\MAD_j(\bv^T \bx_j)}
\end{equation}
where $\mathcal{B}= \{\bv \, ; \|\bv\|=1\}$ is the set of 
all $p$-variate vectors of length 1, $\med$ denotes the 
median, and $\MAD$ is the median absolute deviation, defined as
\begin{equation} \label{eq:mad}
  \text{MAD}(y_1,\ldots,y_n) = 
  1.483\,\text{med}_i| y_i -\text{med}_j(y_j)|
\end{equation}
for a univariate sample $y_1,\ldots,y_n$\,.
The SDO is a projection pursuit method. 
It is based on the idea that if a point
is a multivariate outlier, then there must be some 
one-dimensional projection of the data in which the point 
is a univariate outlier. It further relies on the 
empirical observation that one-dimensional projections of 
many high-dimensional data clouds often have a roughly
normal distribution. 
   
\begin{figure}[ht]
\centering
\includegraphics[width=0.87\linewidth]
{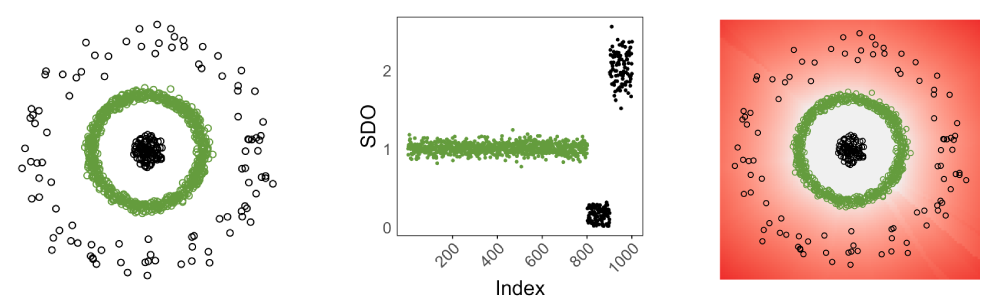}
\caption{Inside-Outside dataset (left), the corresponding 
SDO values (middle) and their heatmap (right).}
\label{fig:insideoutside}
\end{figure}

However, these assumptions are not always satisfied in real 
data. We illustrate the SDO on the bivariate dataset in the 
left panel of Figure~\ref{fig:insideoutside}. We call it the 
`Inside-Outside data set', as it has 800 regular observations 
scattered around a circle (in green), 100 clustered outliers 
inside the circle, and 100 outliers near a circle with a 
larger radius. The outliers are shown in black. The middle 
panel shows the SDO values of these 1000 points. We see that 
the inner cluster has much smaller outlyingness than the 
regular points. 
    
Note that the SDO can also be computed in an arbitrary point
$\bx$, which does not have to be a data point. All we need to
do in \eqref{eq:sdo} is replace $\bx_i$ by $\bx$, yielding
$\text{SDO}(\bx) = \sup_{\bv \in \B} \big\{\left| \bv^T 
\bx-\med_j(\bv^T\bx_j)\right|/\MAD_j(\bv^T\bx_j)\big\}$.
The rightmost panel of Figure~\ref{fig:insideoutside} is a 
heatmap, obtained by computing the SDO on a fine grid of 
points $\bx$ and coloring the results. The region with an 
SDO below the median SDO of the entire dataset is colored 
white, and the color gradually goes from white to red for 
increasing outlyingness. We see that the SDO does not
flag the data points inside the regular circle as 
outliers.

The SDO values in Figure~\ref{fig:insideoutside} were
obtained from the function
\code{depth.projection} in the \proglang{R} 
package \pkg{ddalpha} \citep{Pokotylo:ddalpha}
which computes the exact SDO values.
Whereas the computation of the exact SDO is feasible 
in small dimensions, it becomes too demanding in high 
dimensions due to the enormous set of unit vectors 
$\mathcal{B}$ in \eqref{eq:sdo}. 
\citet{maronna1995behavior} noted that even when
replacing the median and the MAD by smooth estimators 
of location and scale, the function inside
\eqref{eq:sdo} has multiple local maxima, which makes 
gradient-based methods ineffective. 
Approximate algorithms to compute the SDO typically 
rely on using a specific subset of directions
in $\mathcal{B}$. The oldest approach is to draw 
a random subsample of $p$ data points, and to use 
a unit vector orthogonal to the affine hyperplane 
they span \citep{stahel1981robust}. This is then
repeated many times. The advantage of generating 
these directions is that the resulting SDO values
do not change when the data are subjected to a
nonsingular linear transformation. This property
is called affine invariance. However, the 
$p$-subset approach becomes very time consuming 
for high-dimensional data, and cannot be applied 
when there are more dimensions than data points.

A less computationally expensive approach is to
consider directions through two randomly selected
data points \citep{hubert2005robpca,mrfDepth}. The
resulting SDO values are invariant to orthogonal
transformations of the data, which is appropriate 
in for instance the context of principal component 
analysis (PCA).
Another algorithm is based on randomly drawing
directions from the uniform distribution on
the hypersphere, as in \citet{Velasco:RPD}. 
But according to \citet{Dyckerhoff:ProjDepth} this
yields poor performance that degrades with increasing 
dimension. 
 
To relax the elliptical assumption underlying the SDO, \citet{hubert2008outlier} proposed a version of the
SDO adjusted for skewed data, based on a robust measure 
of skewness. A different approach is proposed 
in \citet{tamamori2023kernel}. First the data are 
transformed to a kernel feature space, then their 
dimension is reduced by kernel PCA, and finally their 
SDO is computed using directions drawn randomly from 
the uniform distribution on the unit hypersphere. Kernel 
transformations can be beneficial in this context,
because they make linear separations more feasible,
which improves the performance of SDO. But as the 
resulting Kernel Random Projection Depth (KRPD) 
relies exclusively on these directions, its ability 
to identify outliers is limited, especially in 
high-dimensional scenarios. This will be illustrated 
empirically in Section~\ref{sec:experimentation}. 

Several other kernel-based outlier detection methods have 
been proposed, such as the One-Class Support Vector 
Machine (OCSVM) of \cite{scholkopf1999support} and the
Kernel Minimum Regularized Covariance Determinant (KMRCD) 
method of \cite{schreurs2021outlier}. The widely adopted 
OCSVM method suffers from sensitivity to hyperparameter 
choices, which is especially problematic in unsupervised 
contexts such as outlier detection, where cross validation
is not possible. The KMRCD method combines robust 
covariance estimation and kernel transformations, but 
still relies on elliptical distribution assumptions in the 
kernel feature space, restricting its generality. 

To address these issues we propose a novel Kernel Outlier 
Detection (KOD) approach. It contains two main novelties:
(i) we introduce intuitive and computationally efficient 
directions that naturally arise from kernel transformations, 
and (ii) we propose a robust aggregation mechanism for 
combining multiple direction types, which enhances 
detection accuracy.

Our empirical validation uses synthetic toy datasets
and high-dimensional image datasets. We compare 
KOD with existing unsupervised methods, including OCSVM,
KMRCD, k-Nearest Neighbors, Local Outlier Factor (LOF) 
\citep{breunig2000lof}, and Isolation Forest (IF) 
\citep{liu2008isolation}. Although no single method 
outperforms all others across all scenarios, KOD 
performed consistently among the best, especially on 
highly contaminated datasets.
    
The remainder of this paper is organized as follows. 
In Section~\ref{sec:methodology}, we describe the proposed 
methodology, and then illustrate it on challenging
small datasets in Section~\ref{sec:toy}. In 
Section~\ref{sec:experimentation}, we describe a series of 
experiments on these small datasets and on large-scale 
real-world datasets, to investigate the performance 
against competing methods. Finally, 
Section~\ref{sec:conclusion} concludes with a discussion.
    
\section{The Kernel Outlier Detection Method} 
\label{sec:methodology}
    
The proposed methodology consists of three main steps. First, we map the data into a high-dimensional feature space using kernel functions, and derive a finite-dimensional representation that preserves the essential structure of the data. Next, we adapt the SDO measure for high-dimensional data by constructing computationally feasible subsets of relevant directions. Finally, we combine the results from these subsets and identify outliers.
    
\subsection{Constructing Feature Vectors}
    
We begin by mapping the data into a feature space. Let $\mathcal{X}$ denote the input space, and consider a dataset $\bX = \{ \bx_1, \dots, \bx_n \} \subset \mathcal{X}$. We define a feature mapping $\phi : \mathcal{X} \to \mathcal{F}$, where the feature space $\mathcal{F}$ is a vector space with an inner product $\langle \cdot,\cdot \rangle_\mathcal{F}$ defined on it. 
For each observation $\bx_i$ we call $\phi(\bx_i)$ its feature vector.
A kernel function $k: \mathcal{X} \times \mathcal{X} \to \R$ is defined by
\begin{equation*}
    k(\bx, \bx') = \langle \phi(\bx), \phi(\bx') \rangle_\mathcal{F} \quad \quad 
    \mbox{ for } \bx,\bx' \in \mathcal{X}.
\end{equation*}
We consider only positive semidefinite (PSD) kernel functions to ensure that $k$ defines a valid inner product in the feature space, as guaranteed by Mercer's theorem \citep{scholkopf2001learning}. A symmetric function $k: \mathcal{X} \times \mathcal{X} \rightarrow \mathbb{R}$ is called PSD if for any finite set  $\left\{ \bx_1, \ldots, \bx_n \right\} \subset \mathcal{X}$ and real coefficients $\left\{c_1,\ldots,c_n\right\} \subset \R$ it holds that
\begin{eqnarray*}
   \sum_{i=1}^n \sum_{j=1}^n c_i c_j
   k(\bx_i, \bx_j) \geqslant 0.
\end{eqnarray*}
A commonly used kernel is the Radial Basis Function 
(RBF) defined as
\begin{equation} \label{eq:RBF}
    k(\bx,\bx') = \exp\left(-\frac{\|\bx - \bx'\|^2}{2\sigma^2}\right),
\end{equation}
where $\sigma$ is a tuning parameter. The RBF kernel is 
bounded, which turns out to yield favorable robustness 
properties \citep{Debruyne:InfluentialObs}. To select 
$\sigma$ we employ the \emph{median heuristic} \citep{gretton2012kernel, schreurs2021outlier} given by
\begin{equation} \label{eq:bandwidth}
  \sigma^2 = \med \left\{
	\|\bx_{i} - \bx_{j}\|^2 \; ; \; 
	1 \leqslant i < j \leqslant n
  \right\}\, .\end{equation}
This heuristic assumes that all the input variables are 
measured in the same units. If this assumption does not 
hold we first standardize the data, for instance by
subtracting the columnwise median and dividing by
the columnwise MAD.  
Note that in supervised learning tasks, there is a  
categorical or numerical response variable that one
tries to fit according to a criterion, and then one 
can choose $\sigma$ through cross-validation. But in 
our context of unsupervised outlier detection 
there is no response variable, so cross-validation is 
not applicable.

From a PSD kernel function one constructs an  
$n \times n$ kernel matrix $\mathbf{K}$ with 
entries $k_{ij} = k(\bx_i, \bx_j)$.
This kernel matrix contains all pairwise inner 
products of the feature vectors, so 
$\mathbf{K} = \mathbf{\Phi} \mathbf{\Phi}^T$ 
with $\mathbf{\Phi} = [\phi(\bx_1), \ldots, \phi(\bx_n)]^T$. 
To work efficiently within the feature space, we aim to obtain a finite-dimensional representation of it, even if the full feature space has infinitely many dimensions. First, we restrict ourselves to the subspace spanned by the feature vectors $\phi(\bx_i)$, as it contains all the available information. To compute this subspace, we consider the centered feature vectors 
\begin{equation*}
    \tilde\phi(\bx_i) = \phi(\bx_i) - 
    \frac{1}{n} \sum_{i=1}^n \phi(\bx_i)
\end{equation*}
that span a subspace $\widetilde{\mathcal{F}}$ of
dimension at most $n-1$.
The $n \times n$ centered kernel matrix 
$\widetilde{\mathbf{K}}$ has entries
\begin{eqnarray*}
  \tilde{k}_{ij} = \langle \tilde\phi(\bx_i),
  \tilde\phi(\bx_j) \rangle_\mathcal{F}\,.
\end{eqnarray*}
It can be computed directly from $\mathbf{K}$ as
\begin{align}  \label{eq:centerK}
  \tilde{k}_{ij} & = \left( \phi(\bx_i) - \frac{1}{n} \sum_{\ell=1}^n \phi(\bx_\ell) \right)^T \left( \phi(\bx_j) - \frac{1}{n} \sum_{\ell'=1}^n \phi(\bx_{\ell}') \right)
  \nonumber \\
  &= k_{ij} - \frac{1}{n} \sum_{\ell=1}^n k_{\ell j} - \frac{1}{n} \sum_{\ell'=1}^n k_{i \ell'} + \frac{1}{n^2} \sum_{\ell=1}^n \sum_{\ell'=1}^n k_{\ell \ell'} \nonumber  \\
  &= \left( \mathbf{K} - \mathbf{1}_{nn} \mathbf{K} - \mathbf{K} \mathbf{1}_{nn} + \mathbf{1}_{nn}\mathbf{K} \mathbf{1}_{nn} \right)_{ij} \; ,
\end{align}
where $\mathbf{1}_{nn}$ is the $n \times n$ matrix with 
all entries equal to $\frac{1}{n}$. 
    
Kernel methods typically work with $\mathbf{K}$ or $\widetilde{\mathbf{K}}$ without the need to explicitly know or compute the feature map. This approach is known as the \emph{kernel trick}. In our proposed outlier detection method, we will however construct an $n \times r$ matrix $\mathbf{F}$ (with $r = \text{rank}(\mathbf{K}) \leqslant n-1$) that satisfies 
\begin{eqnarray*}
    \widetilde{\mathbf{K}} = \mathbf{F} \mathbf{F}^T.
\end{eqnarray*}
The matrix $\mathbf{F}$ can easily be derived from the spectral decomposition $\widetilde{\mathbf{K}} = \mathbf{V} \mathbf{\Lambda} \mathbf{V}^T$ of the centered kernel matrix, where $\mathbf{V}$ contains the eigenvectors of $\widetilde{\mathbf{K}}$ corresponding to the $r$ strictly positive eigenvalues $\lambda_1 \geqslant \lambda_2 \geqslant \dots \geqslant \lambda_r > 0$ that form the diagonal matrix $\mathbf{\Lambda}$. The matrix $\mathbf{F} = \mathbf{V}\mathbf{\Lambda}^{1/2}$ then satisfies $\mathbf{F}\mathbf{F}^T = \mathbf{V} \mathbf{\Lambda}^{1/2} (\mathbf{V} \mathbf{\Lambda}^{1/2})^T = \mathbf{V} \mathbf{\Lambda}\mathbf{V}^T = \widetilde{\mathbf{K}}$.

Note that this decomposition is not unique, as for any $r \times r$ orthogonal matrix $\mathbf{U}$ also $\mathbf{F}\mathbf{U}$ satisfies $(\bF \bU)(\bF \bU)^{T} = 
    \bF \bU\bU^{T}\bF^{T} = 
    \bF\bF^{T} = \widetilde{\bK}$.
However, we prefer to work with the matrix $\bF$ because its columns are ranked by decreasing variance.

A different way to compute $\bF$ is by means of a transformation matrix. Since all diagonal entries of $\bL$ are strictly positive, we can compute
$\bL^{-1/2}$ as the $r \times r$ diagonal 
matrix with diagonal entries $1/\sqrt{\lambda_1},\ldots,1/\sqrt{\lambda_r}$\,.
We then construct the $n \times r$ 
transformation matrix $\bT$ as
   $\bT\,:=\, \bV \bL^{-1/2}$.
Then $\bF=\widetilde{\bK}\bT$ since
\begin{equation*}
  \widetilde{\bK} \bT = (\bV \bL \bV^T) (\bV \bL^{-1/2}) =
	\bV \bL (\bV^T \bV) \bL^{-1/2}
	= \bV \bL^{1/2} = \bF\;.
\end{equation*}

The rows $\bef\!_i$ of $\mathbf{F}$ are 
representations of the data points in the 
subspace $\widetilde{\mathcal{F}}$.
They are equal to the feature vectors 
$\tilde{\phi}(\bx_i)$ up to an orthogonal 
transformation. At large sample sizes, they 
can still be high dimensional, making all 
following computations intensive. Hence, in 
the decomposition of $\widetilde{\bK}$, 
we retain only the $q$ largest eigenvalues 
such that their cumulative sum accounts for 
at least 99\% of the total sum, i.e., we set 
$q$ as the smallest value for which 
\begin{equation} \label{eq:explainedvar}
   \sum_{j=1}^q \lambda_j \, \Big/ \,  
   \sum_{j=1}^{r} \lambda_j \geqslant 0.99 \,.
\end{equation} 
We define $\mathbf{\Lambda}_q$ as the $q \times q$ 
diagonal matrix of the retained eigenvalues, as well
as the $n \times q$ matrix $\mathbf{V}_q$ containing 
the corresponding eigenvectors. Then we construct 
the matrix $\hat{\mathbf{F}} = 
\mathbf{V}_q \mathbf{\Lambda}_q^{1/2}$ which 
satisfies $\hat{\mathbf{F}}\hat{\mathbf{F}}^T 
\approx \widetilde{\mathbf{K}}$. 
We can also compute $\hat{\mathbf{F}}$ as
\begin{equation} \label{eq:hatf}
   \hat{\mathbf{F}} = \widetilde{\bK} \bT_q \quad 
   \text{with} \quad \bT_q= \bV_q \bL_q^{-1/2}\;.
\end{equation}
We call the rows $\boldsymbol{\hat{f}}\!_i$  
of $\hat{\mathbf{F}}$ the 
\textit{approximate feature vectors}. 
They can also be obtained as the scores from applying 
PCA to the feature vectors $\phi(\bx_i)$, known as 
kernel PCA \citep{scholkopf1998nonlinear}. 
But whereas kernel PCA typically aims to reduce the
dimension a lot (i.e., using a low $q$), our approach 
focuses on preserving enough structure for outlier 
detection. This motivates selecting the high threshold 
of 0.99 in Equation~\eqref{eq:explainedvar}.
The computation of $\hat{\mathbf{F}}$ can be carried 
out using the \code{makeFV} function in the 
\proglang{R} package \pkg{classmap} 
\citep{raymaekers2022class,classmap}.  

Note that not all kernel matrices originate from
numerical data. For instance, a string kernel can
produce a kernel matrix from text or parts of
DNA. Our methodology can handle such kernel 
matrices as well.

\subsection{Constructing Directions}
    
The second step of the KOD method is to compute
a measure of outlyingness applied to the 
approximate feature vectors $\bhfi$\,. 
In order to compute the SDO of~\eqref{eq:sdo}
we need a set of directions. For this purpose, we 
construct four different types of directions, 
that we will then combine. 

\vspace{2mm}
\noindent {\bf One Point type}: a first set of directions
$\mathcal{B}_1$ consists of the $n$ directions passing 
through each point $\bhfi$ and a robust center. 
Similar to the choices made in \citet{Hubert:RAPCA}
and \citet{croux2005high} in the context of robust PCA, 
we use the $L_1$-median as center. This estimator can 
withstand up to 50\% of outliers. It is also 
orthogonally invariant, which is relevant in our context
since rotations in feature space leave the kernel
matrix unchanged, and the orthogonal equivariance of
this center ensures that the SDO values remain 
unchanged 
as well. The $L_1$-median is computed using the NLM 
algorithm described in \citet{Fritz:L1median} and 
provided in the \proglang{R} package \pkg{pcaPP} 
\citep{filzmoser2006pcapp}.
At large sample sizes $n$, a random subset of 
$\mathcal{B}_1$ could be used, but in our examples we 
will consider all $n$ directions.

\vspace{2mm}    
\noindent {\bf Two Point type}: the second set of 
directions $\mathcal{B}_2$ is given by lines passing 
through pairs of feature points $\bhfi$ and $\bhf\!_j$ 
as in \citet{hubert2005robpca}. We restrict 
$\mathcal{B}_2$ to at most 5000 directions, selected
randomly from the $\binom{n}{2}$ possible directions.

\begin{figure}[ht]
\centering
\includegraphics[width=0.4\linewidth]{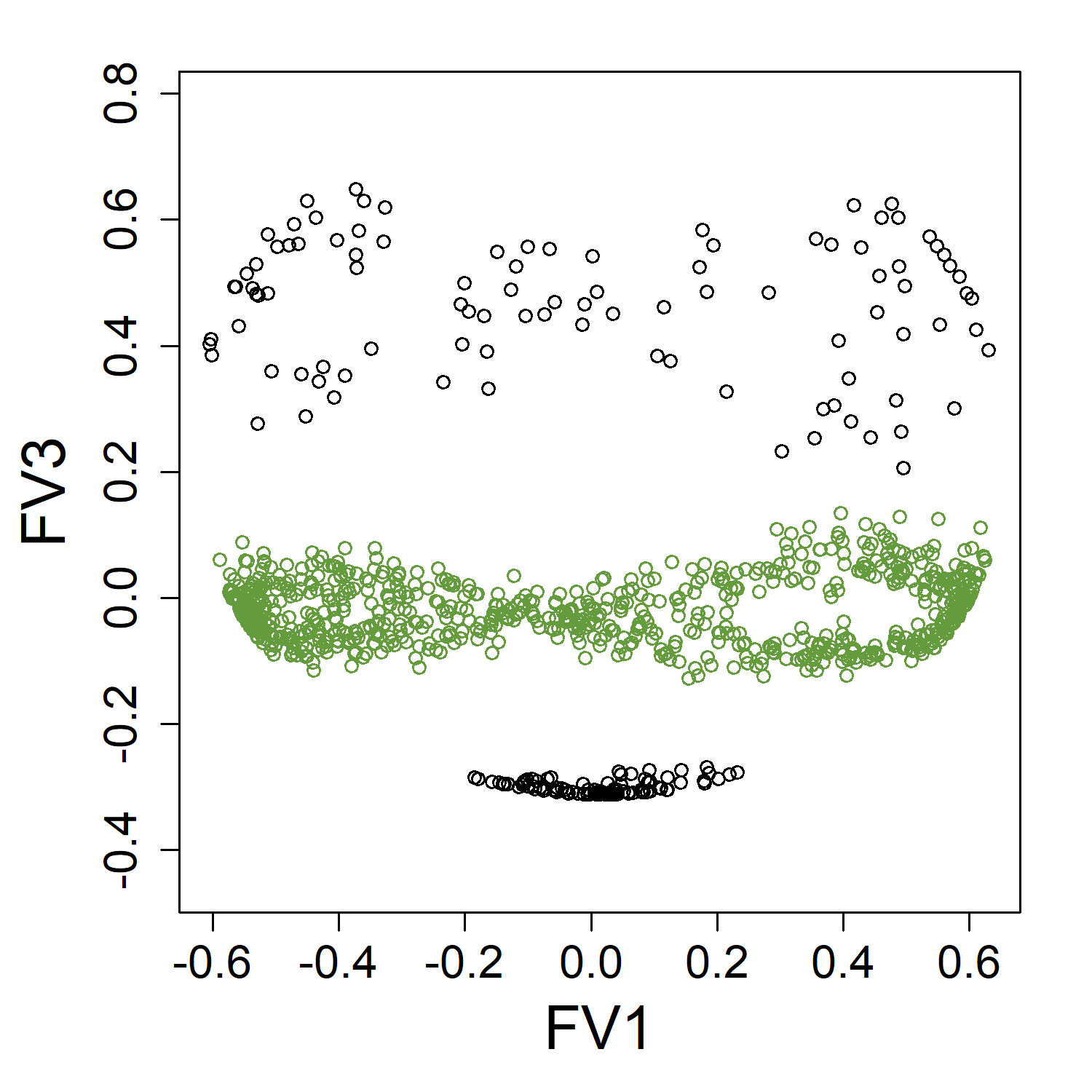}

\vspace{-3mm}
\caption{Scatterplot of the third versus the first 
        coordinate in feature space, of the
        Inside-Outside dataset of
        Figure~\ref{fig:insideoutside}.}
\label{fig:fvpairs}
\end{figure}

\vspace{2mm}    
\noindent {\bf Basis Vector type}: the third set 
$\mathcal{B}_b$ is new. It consists of the $q$ basis 
vectors computed when constructing the $\bhfi$
and is natural in our context. Since they correspond 
to the first $q$ principal components of the feature 
vectors $\phi(\bx_i)$, it can be expected that some of 
these basis vectors point in the direction of outliers 
because the resulting projections yield large variances.
We observed empirically that this often does happen. In Figure~\ref{fig:fvpairs}, we see a scatterplot of the $(\hat{f}_{i1},\hat{f}_{i3})$ for the Inside-Outside 
dataset of Figure~\ref{fig:insideoutside} after applying
the RBF kernel. The kernel transformation has created a 
linear separation between the three groups in the data. 
Projecting the points onto the third basis vector then
yields large SDO values for both the `outside' and
the `inside' outliers. 

\vspace{2mm}
\noindent {\bf Random type}: we also randomly 
draw vectors from the unit hypersphere 
in $q$ dimensions. This set of directions we denote
as $\mathcal{B}_r$. Our default number of random 
directions is set to 1000 as in 
\citet{tamamori2023kernel}.
    
\subsection{Flagging Outliers} \label{sec:flag}

For each set of directions, we can now compute the SDO 
of the $\bhfi$ as in \eqref{eq:sdo}. In experiments, we 
noticed that for some directions the denominator could 
become extremely small, thereby unduly enlarging the
result. To avoid this, we impose a lower bound $c_d$
on the denominator. This $c_d$ is obtained in a 
data-driven manner by
\begin{equation} \label{eq:denom}
    c_d = \frac{1}{5} \med_{\boldsymbol{v} \in 
    \mathcal{B}_r}\left( \text{MAD}_{j}
    (\boldsymbol{v}^T \bhfj) \right)\;.
\end{equation}

Next, we compute the outlyingness for each set of 
directions. The index \texttt{type} ranges over
1, 2, b and r, corresponding to the sets
$\B_1$\,, $\B_2$\,, $\B_b$ and $\B_r$\,.
For each \texttt{type} we get
\begin{equation} \label{eq:outltype}
  \text{outl}_\text{type} (\bhfi) =
  \max_{\boldsymbol{v} \in \mathcal{B}_\text{type}} 
  \,\frac{\left| \bv^T \bhfi - \med_j(\bv^T \bhfj )
  \right|}{ \max\left(\MAD_j(\bv^T \bhfj ),\,  
  c_d\right)}.
\end{equation}
    
Now we still have to combine these four measures of
outlyingness into a single one. For this purpose, we
first normalize each outl\textsubscript{type} by its 
median, and then take the largest normalized
value. This yields the \textit{Kernel Outlyingness} 
(KO) of each case $i$, given by
\begin{equation} \label{eq:ko}
  \text{KO}_i = \max_\text{type} \left( 
  \frac{\text{outl}_\text{type}(\bhfi)}
  {\med_j(\text{outl}_\text{type}(\bhfj)) } \right).
\end{equation}
The normalization ensures that no single direction 
type dominates, as we observed empirically that 
outl\textsubscript{type} can have a rather different 
behavior for different types.
    
The last task is to flag outliers based on the 
$\text{KO}$ values. Empirical evidence, such as 
provided by \citet{rousseeuw2018measure}, suggests 
that in high-dimensional settings the distribution 
of outlyingness values for non-outliers often 
approximates a log-normal distribution. Hence, we 
first apply a logarithmic transformation to 
the KO$_i$ \,:
\begin{equation} \label{eq:log}
    \text{LO}_i = \log(0.1 + \text{KO}_i),
\end{equation}
and then estimate their center and scale with the 
Huber M-estimator of location $\hat{\mu}_M$ 
\citep{Huber:Robloc} and the Q\textsubscript{$n$}
estimator of scale $\hat{\sigma}_{Qn}$ 
\citep{Rousseeuw:Scale}. 
Next, we determine the cutoff value as
\begin{equation} \label{eq:cutoff}
  c = \exp\big( \hat\mu_M( \text{LO}) + 
  z_{0.99} \hat\sigma_{Qn}( \text{LO} )
  \big) - 0.1,
\end{equation}
where $z_{0.99}$ is the 99th percentile of the 
standard normal distribution. Finally, the data 
points with $\text{KO}_i \geqslant c$ are 
flagged as outliers.
    
The full procedure is summarized in
Algorithm~1.
\begin{algorithm}
\caption{Kernel Outlier Detection}
\begin{algorithmic}[1]
\Require Dataset $\bX = \{\bx_1, \dots, \bx_n \} \subset \mathcal{X}$, kernel function $k: \mathcal{X} \times \mathcal{X} \to \R$
    
\State Compute the kernel matrix $\mathbf{K}$,
and its centered version $\widetilde{\mathbf{K}}$
given by~\eqref{eq:centerK}.   
\State Perform the spectral decomposition of $\widetilde{\mathbf{K}}=\mathbf{V}\mathbf{\Lambda}\mathbf{V}^T $ and construct the approximate feature matrix $\hat{\mathbf{F}} = \mathbf{V}_q \mathbf{\Lambda}_q^{1/2}$\,.
    
\State Create the subsets of directions $\mathcal{B}_1$\,, $\mathcal{B}_2$\,, $\mathcal{B}_b$ and $\mathcal{B}_r$\,.
    
\State Compute a lower bound for the denominator according to \eqref{eq:denom}.
    
\State Compute the outlyingness for each $\bhfi$ and each direction type, following \eqref{eq:outltype}.
    
\State Scale the outlyingness values by their medians and compute the kernel outlyingness as in \eqref{eq:ko}.
    
\State Obtain a cutoff $c$ for outlier detection using \eqref{eq:log} and \eqref{eq:cutoff}.
    
\State Flag data points as outliers if $\text{KO}_i \geqslant c$.
    
\Ensure Kernel outlyingness $\text{KO}_i$ of all cases, 
and the set of flagged outliers.
\end{algorithmic}
\end{algorithm}
    
When we need to compute the KO of out-of-sample 
cases $\{\by_1,\dots,\by_m\}$, we first compute 
all the kernel values $k^{yx}(\by_i,\bx_j)$
for $i=1,\dots,m$ and $j=1,\dots,n$ and store them 
in the $m \times n$ matrix $\bK^{yx}$. Then we 
compute the centered kernel matrix 
\begin{equation*}
\widetilde{\bK}^{yx} = \bK^{yx} - 
  \bK^{yx}\mathbf{1}_{nn} - 
  \mathbf{1}_{mn} \bK + 
  \mathbf{1}_{mn} \bK \mathbf{1}_{nn}
\end{equation*}
with $\mathbf{1}_{mn}$ the $m \times n$ matrix 
with all elements equal to $1/n$. Finally we 
compute the approximate feature vectors  
$\hat{\bF}^y = \widetilde{\bK}^{yx} \bT_q$ as 
in~\eqref{eq:hatf}. For each set of directions
$\mathcal{B}_\text{type}$\,, the outlyingness 
of each $\hat{\bef^y}$ is computed as 
in~\eqref{eq:outltype} where the center 
$\med_j(\bv^T \bhfj)$ and scale 
$\MAD_j(\bv^T \bhfj)$ in each direction are
those from the training data.

\section{Toy datasets}\label{sec:toy}

To support the rationale behind the proposed KOD 
method and provide an intuitive understanding of 
its strengths, we first show detailed results 
on three synthetic but challenging bivariate 
datasets. These `toy' datasets illustrate how 
different types of projection directions 
influence outlier detection performance.

\begin{figure}[ht]
\centering
\includegraphics[width=0.96\linewidth]
  {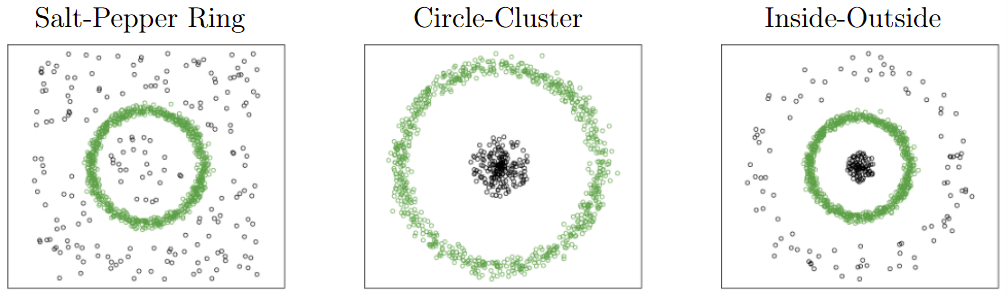}
\vspace{-1mm}
\caption{Scatter plots of three toy datasets. Green 
dots represent regular data points, and black dots 
represent outliers.}
\label{fig:toydatasets}
\end{figure}

The three datasets are displayed in 
Figure~\ref{fig:toydatasets} and referred to as 
\emph{Salt-Pepper Ring}, \emph{Circle-Cluster},
and \emph{Inside-Outside}. 

\begin{itemize}
\item \textbf{Salt-Pepper Ring:} In this dataset, 
the regular data points are arranged in a roughly 
circular pattern, while the outliers are uniformly 
sprinkled both inside the circle and in the 
background, creating sparse noise with no 
coherent structure.
\item \textbf{Circle-Cluster:} The regular data 
points again follow a circular pattern, but the 
outliers now form a cluster near the center. 
This is harder than the Salt-Pepper Ring, since
some methods may interpret the centrally located 
outliers as a dense cluster of regular points.
\item \textbf{Inside-Outside:} This dataset was 
already shown in Section~\ref{sec:introduction}. 
It contains two concentric circular patterns and 
a cluster in the center. This is the most
challenging setup because it combines clustered 
outliers at the center with rather structured
outliers outside the main circle. We put the
same number of outliers outside as inside.
\end{itemize}
Each dataset contains $n = 1000$ observations, 
with contamination levels set at 20\%.
In Section~\ref{sec:experimentation}, we will 
compare KOD with competing methods on these 
datasets, and then we will use contamination 
levels of 5\%, 10\%, and 20\%. 

We apply the RBF kernel to each of these datasets.
They all have 1000 cases so the kernel matrix
is always $1000 \times 1000$, but the centered 
kernel matrices are of ranks 98, 61 and 79 (we 
discard eigenvalues below $10^{-12}$). 
Following~\eqref{eq:explainedvar} the retained number 
of dimensions $q$ is 9, 6, and 8. So even though we 
have applied the RBF kernel, our computations are 
performed in a space of dimension much lower
than $n-1 = 999$.

\begin{figure}[!ht]
\centering
\vspace{2mm}
\includegraphics[width=1\linewidth]{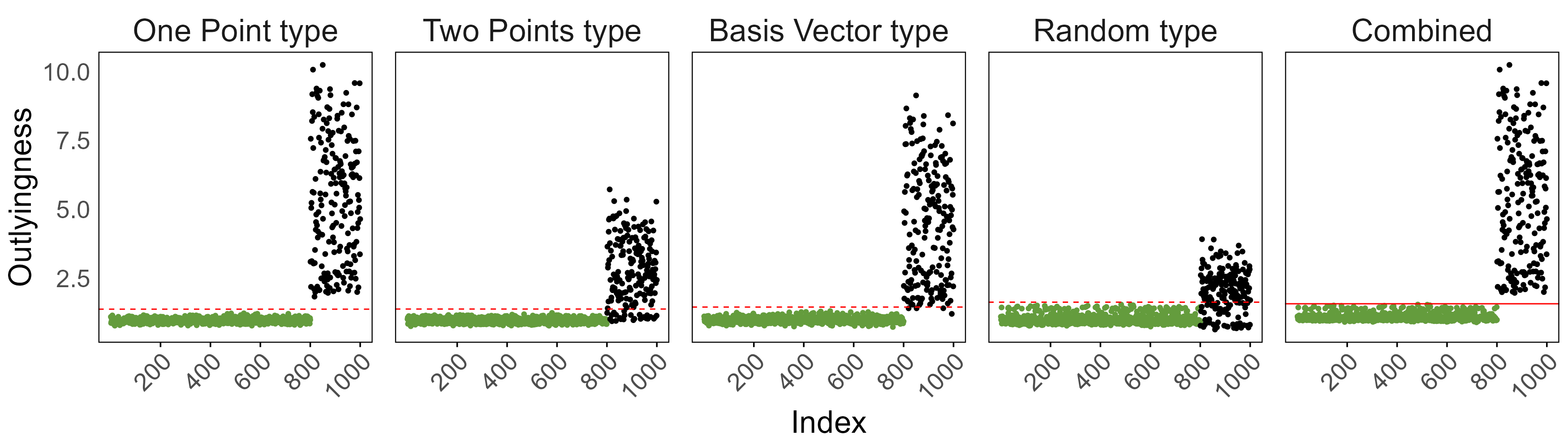}\\
\vspace{-2mm}
\caption{Outlyingness values obtained from each 
direction type and the combined KO outlyingness 
on the \textit{Salt-Pepper Ring} dataset with 
20\% contamination.}
\label{fig:saltycirc_outl1}
\end{figure}

\begin{figure}[!ht]
\centering
\vspace{2mm}
\includegraphics[width=1\linewidth]
{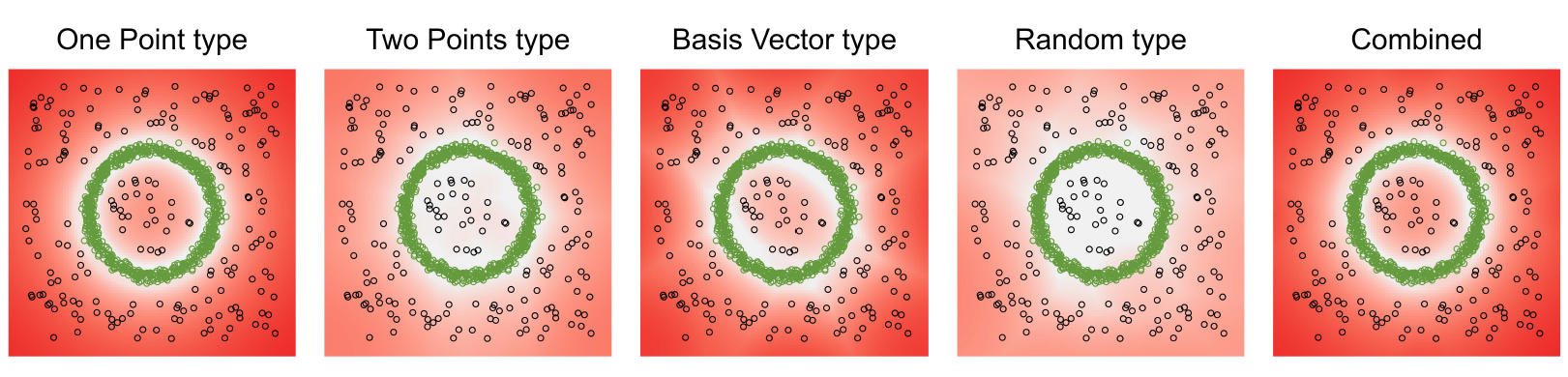}\\
\vspace{-3mm}
\caption{Heatmaps of the outlyingness values 
obtained from each direction type and the 
combined KO on the \textit{Salt-Pepper Ring} 
dataset with 20\% contamination.}
\label{fig:saltycirc_grid1}
\end{figure}

Figures~\ref{fig:saltycirc_outl1} 
and~\ref{fig:saltycirc_grid1} show some results 
on the \textit{Salt-Pepper Ring} dataset. The 
first four panels display the normalized 
outlyingness values $\text{outl}_\text{type}/
\med_j(\text{outl}_\text{type}(\bhfj))$ for 
each type of direction. The last panel shows 
their maximum, which is the final KO. To each 
panel we have added a horizontal line. 
The solid red line in the last panel has height 
$c$, the cutoff given by~\eqref{eq:cutoff} for 
flagging the outliers. The dashed red lines in 
the other panels represent the hypothetical 
thresholds that would be obtained if only that 
single direction type was used for deriving an 
outlier cutoff.

In Figure~\ref{fig:saltycirc_outl1}, the outliers 
were put at the end for visual support, but in 
real data they can be anywhere. 
The Two Points type and the Random type directions 
detected most of the outliers outside the circle, 
but were unable to discover the centrally located 
anomalies. They consider the entire region inside 
the circle as a regular region. The Basis Vector 
type roughly captures the shape of the ring, but 
does not detect all outliers. The One Point type 
did provide a clear separation between regular 
data points and outliers, which is also reflected 
in the combined KO values in the rightmost panel.

The differences between the direction types can be 
seen more clearly in Figure~\ref{fig:saltycirc_grid1},
which displays heatmaps of the SDO computed 
on a grid as in Figure~\ref{fig:insideoutside}. 
Again points whose outl$_\text{type}$ is below
$\med_j(\text{outl}_\text{type}(\bhfj))$ are colored 
white, and darkening shades of red indicate 
increasing outlyingness. The Two Points and Random
types leave the region inside the circle almost 
white, so they failed to capture the structure of 
the regular region. The heatmap of the combined 
directions in the rightmost panel looks a lot 
like that of the One Point type, which in this 
dataset gave the best separation.

\begin{figure}[!ht]
\centering
\vspace{2mm}
\includegraphics[width=1\linewidth]{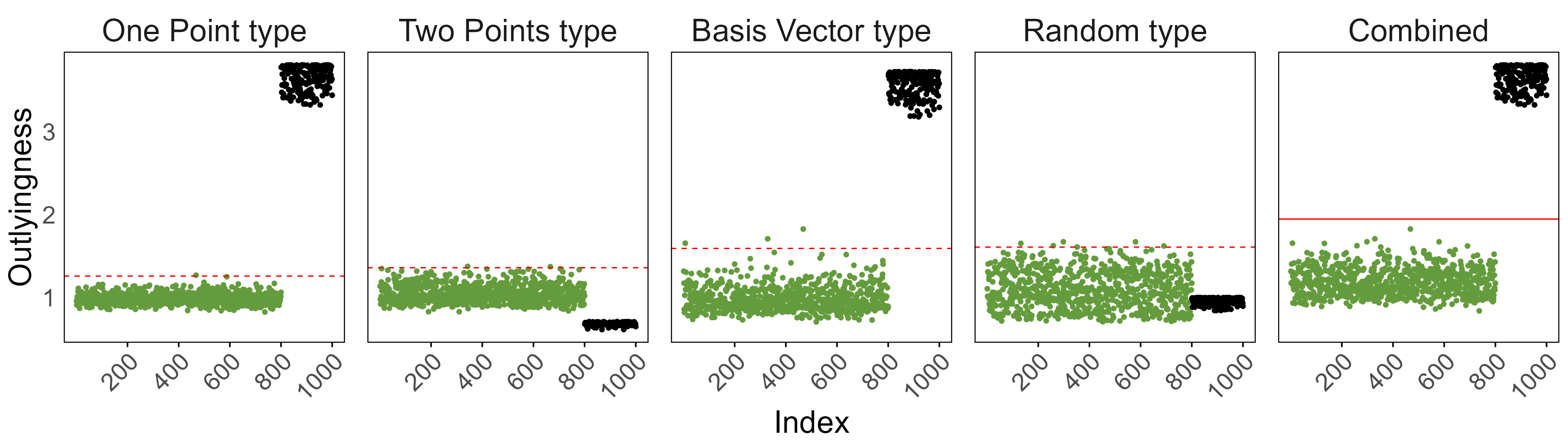}\\
\vspace{-2mm}
\caption{Outlyingness values obtained from each 
direction type and the combined KO outlyingness 
on the \textit{Circle-Cluster} dataset 
with 20\% contamination.}
\label{fig:circlust_outl1}
\end{figure}

\begin{figure}[!ht]
\centering
\vspace{2mm}
\includegraphics[width=1\linewidth]
  {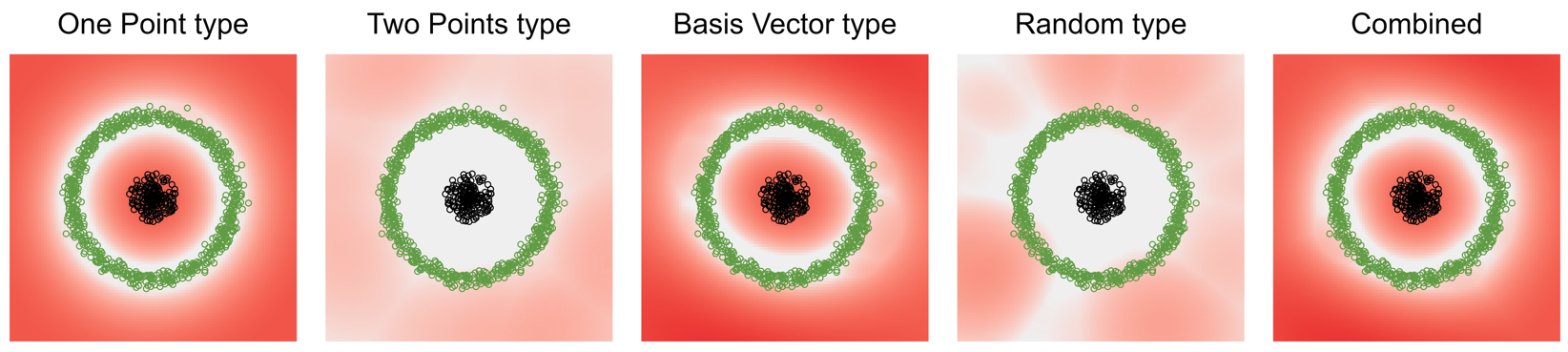}\\
\vspace{-3mm}
\caption{Heatmaps of the outlyingness values 
obtained from each direction type and the 
combined KO on the \textit{Circle-Cluster} 
dataset with 20\% contamination.}
\label{fig:circlust_grid1}
\end{figure}

Figures~\ref{fig:circlust_outl1} and 
\ref{fig:circlust_grid1} show the results on 
the \textit{Circle-Cluster} dataset. The Two 
Points type directions and the Random type 
directions were unable to discover the centrally 
located anomalies. They consider the entire 
region inside the circle as a regular region. 
The One Point and Basis Vector types did provide 
a clear separation between regular data points 
and outliers.

\begin{figure}[!ht]
\centering
\vspace{2mm}
\includegraphics[width=1\linewidth]
  {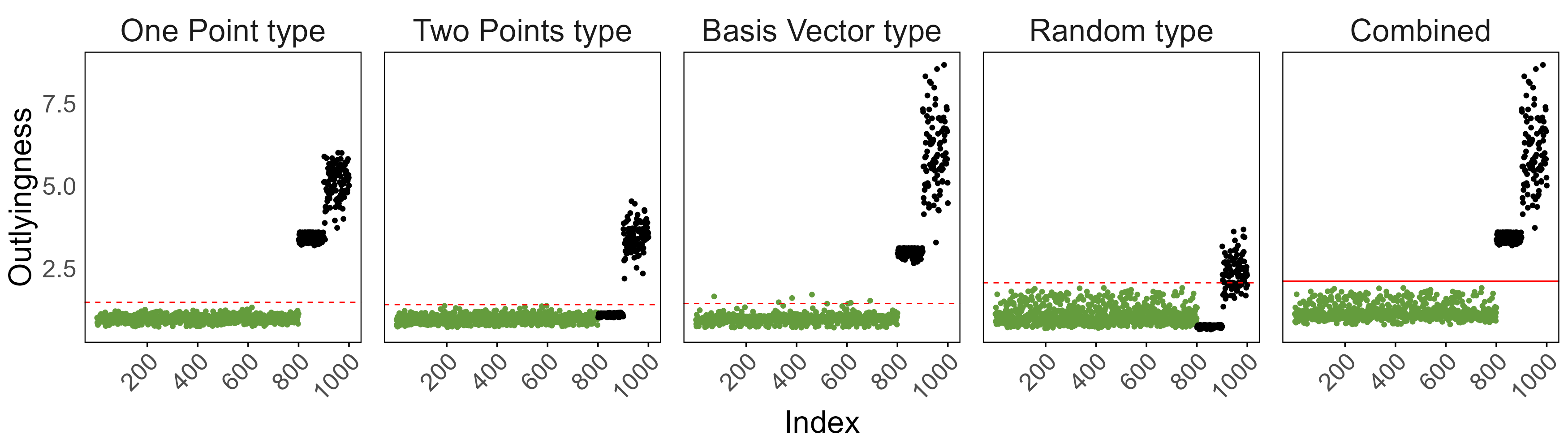}\\
\vspace{-2mm}
\caption{Outlyingness values obtained from each 
direction type and the combined KO outlyingness 
on the \textit{Inside-Outside} dataset with 
20\% contamination.}
\label{fig:twocirc_outl1}
\end{figure}

\begin{figure}[!ht]
\centering
\vspace{2mm}
\includegraphics[width=1\linewidth]
  {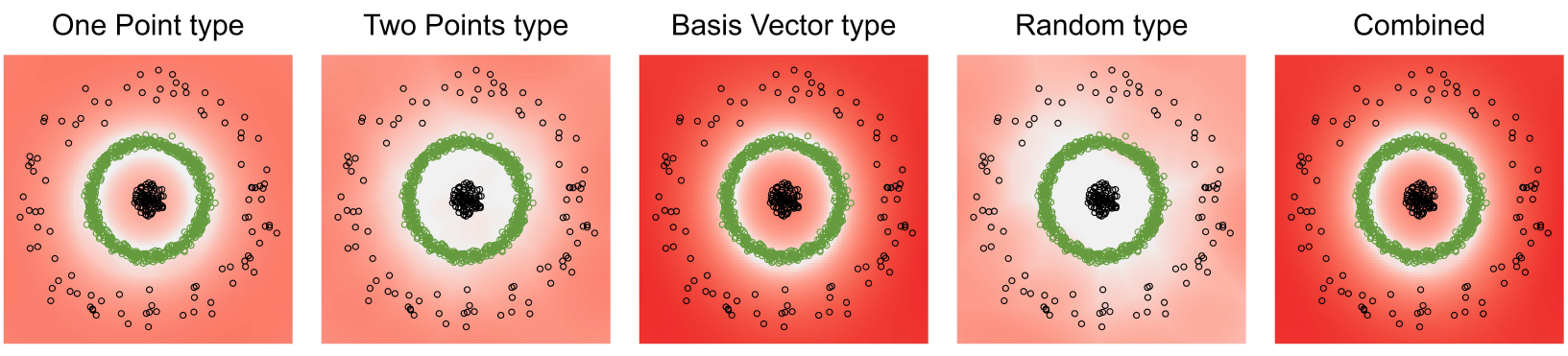}\\
\vspace{-3mm}
\caption{Heatmaps of the outlyingness values 
obtained from each direction type and the 
combined KO on the \textit{Inside-Outside} 
dataset with 20\% contamination.}
\label{fig:twocirc_grid1}
\end{figure}

We now revisit the \textit{Inside-Outside} dataset
of Figure~\ref{fig:insideoutside}. Both the `inside' 
and `outside' outliers need to be identified, which 
SDO cannot do in the input space as  
illustrated in Section~\ref{sec:introduction}.
In Figures~\ref{fig:twocirc_outl1} 
and~\ref{fig:twocirc_grid1}, we see that Two Points
and Random struggle to separate the `inside' 
outliers from the regular observations, whereas
One Point and especially Basis Vector correctly 
identify the region of the regular points. The 
cutoff on the combined KO provides a good 
separation.

In the Appendix, an additional
toy dataset is studied.

\section{Numerical experiments} 
\label{sec:experimentation}

In this section, we compare KOD with several competitors, 
both on the toy datasets of Section~\ref{sec:toy} and 
on four large benchmark datasets. The choice of datasets
was guided by the principle that in order to evaluate
outlier detection methods we need to know which cases
are the outliers, which does not happen very often in real 
data. We will first describe the competing methods and 
the evaluation metrics. 

\subsection{Competing Methods} \label{sec:competingmethods}

We compare KOD with three other kernel-based methods, 
as well as three well-known anomaly detection methods
that do not use kernels.

\vspace{2mm}
\noindent {\bf Kernel random projection depth (KRPD).} \citet{tamamori2023kernel} first applies the kernel,
and then switches to the kernel PCA scores of the 
data points. Next, 1000 directions are generated 
randomly from the uniform distribution on the hypersphere, 
from which the SDO of \eqref{eq:sdo} is computed.
Also a monotone decreasing function of the SDO is
considered, given by 
$D(\bx_i) := 1/(1 + \SDO(\bx_i))$ and called 
projection depth. The paper does not provide an 
unsupervised selection method for the $\sigma$ of 
the RBF kernel, or for the number of principal 
components. 
To make a fair comparison we use the same number of 
principal components as in KOD, and consider 8000 
random directions, which is roughly the total number 
of directions in KOD when $n=1000$.

\vspace{2mm}
\noindent \textbf{Kernel Minimum Regularized Covariance Matrix (KMRCD).} The KMRCD method of \citet{schreurs2021outlier} computes a robust regularized covariance estimator in the feature space. The method uses a fast algorithm that exploits the kernel trick to speed up computations. The hyperparameter $\alpha$ that determines the proportion of data points used to estimate the covariance matrix is set to $0.5$ for maximal robustness. The resulting outlier diagnostic is the robust Mahalanobis distance computed in the feature space. We have implemented this method 
in \proglang{R}.

\vspace{2mm}
\noindent \textbf{One-Class Support Vector Machine (OCSVM).} OCSVM maps the data into the feature space and searches for a hyperplane that  separates them from the origin with maximum margin \citep{scholkopf1999support}. It needs an additional hyperparameter $\nu$ that controls the maximum allowed fraction of  outliers. We set it to 0.5 for maximal robustness. Moreover, OCSVM provides a decision function whose values indicate the position of each point relative to the separating boundary. These decision values can thus be interpreted as an outlyingness values. Computations are performed with the \code{svm} function from the \proglang{R} package \pkg{e1071} \citep{e1071}.

\vspace{2mm}
\noindent \textbf{$\mathbf{k}$NN for outlier detection (KNN).} 
This method and the following ones do not use kernels. KNN ranks the observations based on the  distance to their $k$-th nearest neighbor \citep{ramaswamy2000efficient}. We use the $L_1$ distance which is more informative than the Euclidean distance on high dimensional data, and we consider both $\sqrt{n}$ and $\log{(n)}$ for the hyperparameter $k$.
Computations are performed with the \code{kNNdist} function from the \proglang{R} package \pkg{dbscan} \citep{dbscan}.

\vspace{2mm}
\noindent \textbf{Local Outlier Factor (LOF).} LOF is a density-based method that assigns to each point an outlyingness value based on how much it deviates from its local neighborhood in terms of density \citep{breunig2000lof}. 
We again used the $L_1$ distance for finding the nearest neighbors, and considered two alternatives for the size of the neighborhoods. First, we set $k=20$. Then, as a hyperparameter-free alternative and as advocated in \citet{breunig2000lof}, we calculated outlyingness values for $k = \{10, 20, 30, 40, 50\}$ and set for each data point its final outlyingness as the maximum among the five values (denoted as LOF Pmax). We use the \code{lof} function from the \proglang{R} package \pkg{dbscan} \citep{dbscan}.

\vspace{2mm} 
\noindent \textbf{Isolation Forest (IForest).} IForest is a tree-based method that recursively partitions the data to isolate outliers \citep{liu2008isolation}. The anomaly score of a point is related to the number of splittings required to isolate it. The results depend highly on hyperparameter choices, and there are numerous alternatives to choose from. We used the default settings from scikit-learn \citep{Pedregosa:scikit-learn} and also considered a density-based alternative. The function \code{isolation.forest} from the \proglang{R} package \pkg{isotree} \citep{isotree} was used. 

\subsection{Evaluation Metrics}\label{sec:eval}

In the literature, one often evaluates outlier detection 
performance by the Area Under the Curve (AUC) of the 
Receiver Operating Characteristic (ROC). However, \citet{campos2016evaluation} show that it is not optimal
for this purpose. 
Another measure is the Matthews Correlation Coefficient 
(MCC), which considers all aspects of the confusion matrix:
the number of true positives (TP), true negatives (TN),
false positives (FP), and false negatives (FN). It is
given by
\begin{equation} \label{eq:MCC}
 \mbox{MCC} = \frac{\mbox{TP} \times \mbox{TN} - 
 \mbox{FP} \times \mbox{FN}}
 {\sqrt{(\mbox{TP} + \mbox{FP})(\mbox{TP} + \mbox{FN})
 (\mbox{TN} + \mbox{FP})(\mbox{TN} + \mbox{FN})}}\,.
\end{equation}
However, when comparing a broad range of methods, 
a practical difficulty arises. Many outlier detection 
methods in the literature output a ranking or score 
indicating the degree of outlyingness, but do not provide 
a cutoff threshold beyond which a data point is flagged
as an outlier. Introducing an arbitrary cutoff for such 
methods could introduce bias, and lead to an inaccurate 
assessment of their performance. We will therefore use a
different evaluation metric, called {\it Precision at N}
and denoted P@N. If N is the known number of true 
outliers, then P@N is defined as
\begin{equation}\label{eq:P@N}
\mbox{P@N} = \frac{\mbox{number of true outliers among 
the N highest values of the criterion}}{\mbox{N}}\,.
\end{equation} 
The P@N measure is particularly well-suited in our 
situation, as it circumvents the need for a threshold,
by focusing on the proportion of true outliers among 
the N top ranked cases. 
For a comprehensive discussion on the 
appropriateness of P@N in outlier detection see
\citet{campos2016evaluation}.

\subsection{Performance comparison on the toy datasets}

Applying KOD and the outlier detection methods listed in
Section~\ref{sec:competingmethods} to the three toy
datasets in Section~\ref{sec:toy} yielded the P@N
results in Table \ref{tab:methods_comparison_TOY}. 
In all three datasets we varied the contamination 
percentage from 5\% to 20\%, and the reported P@N is
the average over 10 replications of each setting.

\begin{table}[ht]
\caption{Comparison of outlier detection methods
    applied to the three toy datasets, with varying
    contamination percentage. The entries are the 
    Precision at N (P@N) of \eqref{eq:P@N}, 
    averaged over 10 replications of each dataset.}
    \label{tab:methods_comparison_TOY}
\begin{center}
\renewcommand{\arraystretch}{0.8}
\begin{adjustbox}{width=\textwidth,center}
\begin{tabular}{lccccccccc}
\toprule
  & \multicolumn{3}{c}{\textbf{Salt-Pepper Ring}} & 
    \multicolumn{3}{c}{\textbf{Circle-Cluster}} & 
    \multicolumn{3}{c}{\textbf{Inside-Outside}} \\
   \cmidrule(lr){2-4} \cmidrule(lr){5-7} \cmidrule(lr){8-10}
  & \multicolumn{3}{c}{Contamination} & \multicolumn{3}{c}{Contamination} & \multicolumn{3}{c}{Contamination} \\
  & 5\% & 10\% & 20\% & 5\% & 10\% & 20\% & 5\% & 10\% & 20\% \\
\midrule
        
\textbf{KOD} & 1.00 & 1.00 & 0.94 & 1.00 & 1.00 & 1.00 & 1.00 & 1.00 & 1.00 \\
        
\textbf{KRPD} & 0.78 & 0.77 & 0.81 & 0.00 & 0.00 & 0.00 & 0.47 & 0.50 & 0.54 \\

\textbf{KMRCD} & 0.39 & 0.46 & 0.60 & 0.00 & 0.00 & 0.00 & 0.22 & 0.34 & 0.41 \\
        
\textbf{OCSVM} & 0.87 & 0.86 & 0.88 & 0.00 & 0.00 & 0.00 & 0.50 & 0.50 & 0.50 \\
        
\textbf{KNN} $k = \sqrt{n}$ & 1.00 & 1.00 & 1.00 & 0.83 & 0.41 & 0.20 & 1.00 & 0.92 & 0.69 \\
\textbf{KNN} $k = \log{(n)}$ & 1.00 & 1.00 & 1.00 & 0.50 & 0.39 & 0.23 & 0.80 & 0.77 & 0.70 \\

\textbf{LOF} $k=20$ & 1.00 & 0.97 & 0.75 & 0.30 & 0.52 & 0.61 & 0.50 & 0.66 & 0.78 \\
\textbf{LOF} Pmax & 1.00 & 1.00 & 0.98 & 0.27 & 0.46 & 0.66 & 0.99 & 0.62 & 0.77 \\

\textbf{IForest} Default & 0.90 & 0.94 & 0.96 & 0.12 & 0.03 & 0.00 & 0.62 & 0.58 & 0.50 \\
\textbf{IForest} Density & 0.98 & 0.98 & 0.99 & 0.98 & 0.60 & 0.14 & 1.00 & 0.92 & 0.58 \\
\bottomrule
\end{tabular}
\end{adjustbox}
\end{center}
\end{table}

We observe that KOD worked well on all three datasets,
even for 20\% of contamination. The performance of the
competing methods was quite variable, ranging from
reasonable to poor. On some datasets, certain methods
failed entirely. The kNN method with $k = \sqrt{n}$ and
the density version of IForest had good performance on
\textit{Circle-Cluster} for 5\% of contamination, but 
degraded substantially as the contamination level 
increased. 

The fact that KOD outperformed KRPD illustrates that
only using directions generated from the uniform
distribution on the hypersphere is not enough. The
combination of direction types in KOD makes it more
versatile. We know from partial results 
that sometimes one direction type does better, and 
sometimes another. We also observed 
that some of the competing methods are 
highly sensitive to hyperparameter choices, and in this
unsupervised setting those cannot be tuned by cross 
validation. Even when hyperparameters are set using prior 
knowledge about the data, no method consistently 
produces satisfactory results. This is because the clean 
data can have many different distributions, and the 
outliers can take many forms.

\subsection{Performance comparison on real datasets} \label{sec:experimentation_mnist}
Now we move on to more complex real-world scenarios.
To evaluate the effectiveness of the proposed KOD method, 
we conducted a series of experiments on three well-known 
benchmark datasets of images. As these datasets contain 
nonlinear relationships and are high-dimensional, they 
provide a suitable testing ground. Also, it is easy for 
the human eye to verify outliers in images. 

\begin{figure}[ht]
\centering
\includegraphics[width=\linewidth]
  {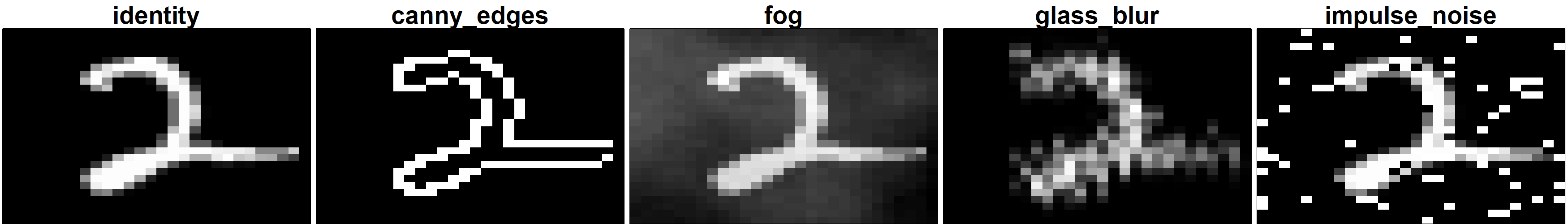}\\
\vspace{-1mm}
\caption{An original image and some 
corruption types in the MNIST-C dataset.}
\label{fig:minstc1}
\end{figure}

The MNIST, MNIST-C, and Fashion MNIST datasets contain 
grayscale images with a resolution of $28 \times 28$ pixels, 
each categorized into 10 distinct classes. The MNIST data 
\citep{LeCunMNIST} consist of handwritten digits from 0 to 9. 
MNIST-C extends the MNIST dataset by introducing several types 
of corruption applied to the handwritten digit images, some of
which are shown in Figure \ref{fig:minstc1}. The Fashion 
MNIST grayscale images depict various types of clothing, 
distributed across 10 categories. 

For MNIST and Fashion MNIST we consider ten different 
setups. In each setup, one of the ten classes was 
designated as the regular images. Next, outliers 
were generated by randomly sampling images from the 
remaining classes. The MNIST-C data already contained
outliers. For each digit and for each corruption type, 
the original digit images were treated as non-outliers, 
while the corresponding corrupted versions were 
considered as outliers. This resulted in a total of 
140 different experimental setups (10 digits by 
14 corruption types).

In addition to these image datasets we also consider
a dataset from \citet{campos2016evaluation}, the
PageBlocks data. It is about blocks in document
pages, and has continuous variables. The regular
data correspond to blocks of text, and the blocks 
with images are considered as outliers.

To investigate the performance under varying outlier 
percentages, we considered three contamination rates: 
5\%, 10\% and 20\%. In all experiments we set the 
sample size $n$ to 1000, and the results were
averaged over 5 replications.

An issue with experiments of this type is that for
real data there is no absolute `ground truth'.
In the image datasets, we observed that some of the
supposedly clean images looked very strange,
and some of the methods indeed flagged them as
outlying. On the other hand, some of the supposedly
contaminated images looked as if they could belong
to the main class, explaining why they were not 
always flagged. This raises the important issue 
of mislabeling in classification, which is an active 
research topic, see e.g. \cite{raymaekers2022class}
and the references cited therein.

For some other datasets,
such as the PageBlocks data, there is a semantic 
justification why some cases are outlying, but is
it really certain that the variables that have been 
measured give enough information about the
difference between clean and outlying cases?
For instance, medical datasets that contain many
healthy people and some sick persons may have
recorded many variables, but do they really 
discriminate between healthy and sick? And 
might not the healthy set contain people 
that are outlying for other reasons?
Moreover, the diagnosis of the study participants 
may be subject to mislabeling also.

The choice of kernel significantly impacts the performance 
of kernel-based methods, as it determines the feature 
space in which outlyingness is computed. 
We used the RBF kernel for all image datasets, with the  
parameter $\sigma$ selected by the median heuristic given 
in \eqref{eq:bandwidth}. The RBF kernel is a reasonable 
choice due to its demonstrated success on image datasets. 
For the PageBlocks dataset, we used 
the linear kernel, but using the RBF kernel did not 
substantially change the performance of KOD.

\begin{table}[!ht]
\caption{Comparison of outlier detection methods applied 
to four real datasets, with varying contamination
percentage. The entries are the averaged P@N, normalized 
by its highest value per column.}
\label{tab:methods_comparison_image}
\begin{center}
\renewcommand{\arraystretch}{0.8} 
\begin{adjustbox}{width=\textwidth,center}
\begin{tabular}{lcccccccccccc}
\toprule
  & \multicolumn{3}{c}{\textbf{MNIST}} & \multicolumn{3}{c}{\textbf{MNIST-C}} & \multicolumn{3}{c}{\textbf{Fashion MNIST}} & \multicolumn{3}{c}{\textbf{PageBlocks}} \\
  \cmidrule(lr){2-4} \cmidrule(lr){5-7} \cmidrule(lr){8-10} \cmidrule(lr){11-13}
  & \multicolumn{3}{c}{Contamination} & \multicolumn{3}{c}{Contamination} & \multicolumn{3}{c}{Contamination}  & \multicolumn{3}{c}{Contamination} \\
  & 5\% & 10\% & 20\% & 5\% & 10\% & 20\% & 5\% & 10\% & 20\% & 5\% & 10\% & 20\% \\
\midrule
        
\textbf{KOD}   & 0.80 & 0.92 & 0.95 & 0.80 & 0.80 & 0.83 & 0.80 & 0.90 & 0.97 & 0.60 & 0.66 & 0.88 \\
        
\textbf{KRPD}  & 0.64 & 0.80 & 0.87 & 0.66 & 0.66 & 0.71 & 0.89 & 0.92 & 0.94 & 0.68 & 0.73 & 0.92 \\

\textbf{KMRCD} & 0.68 & 0.92 & 1.00 & 0.66 & 0.66 & 0.72 & 0.89 & 0.95 & 0.99 & 0.41 & 0.55 & 0.83 \\
        
\textbf{OCSVM} & 0.68 & 0.83 & 0.84 & 0.69 & 0.69 & 0.68 & 1.00 & 0.98 & 0.96 & 0.40 & 0.43 & 0.55 \\
        
\textbf{KNN}  $k = \sqrt{n}$     & 0.78 & 0.92 & 0.94 & 0.78 & 0.76 & 0.77 & 0.96 & 0.97 & 0.94 & 0.73 & 0.79 & 0.84 \\
\textbf{KNN}  $k = \log{n}$      & 0.85 & 0.95 & 0.94 & 0.69 & 0.73 & 0.76 & 0.77 & 0.73 & 0.75 & 0.76 & 0.80 & 0.81 \\

\textbf{LOF} $k=20$     & 0.97 & 0.93 & 0.87 & 0.59 & 0.62 & 0.65 & 0.48 & 0.35 & 0.38 & 0.82 & 0.91 & 0.91 \\
\textbf{LOF} Pmax               & 1.00 & 1.00 & 0.95 & 0.66 & 0.68 & 0.72 & 0.68 & 0.54 & 0.46 & 1.00 & 1.00 & 1.00 \\

\textbf{IForest} Default   & 0.64 & 0.78 & 0.83 & 0.94 & 0.94 & 0.96 & 0.98 & 1.00 & 1.00 &  0.68 & 0.73 & 0.85 \\
\textbf{IForest} Density   & 0.61 & 0.75 & 0.76 & 1.00 & 1.00 & 1.00 & 0.75 & 0.78 & 0.79 & 0.51 & 0.60 & 0.65 \\
\bottomrule
\end{tabular}
\end{adjustbox}
\end{center}
\end{table}

Table \ref{tab:methods_comparison_image} shows the performance 
of KOD and the competing methods in terms of P@N, which has
been normalized by dividing each P@N entry by the highest P@N 
in its column. The performance of KOD was fairly reliable. The 
performance of some other methods was more variable across 
settings. Note that KOD does not require tuning of 
hyperparameters, making it easy to apply. 

The Appendix contains an additional table in 
which the performance is measured by the Matthews correlation
of~\eqref{eq:MCC} instead of P@N. There KOD is only compared 
to the three other methods that provide cutoff values. The
results are qualitatively similar. Also a figure with
computation times is provided there.

\section{Conclusion}\label{sec:conclusion}

In this paper, we introduced a new anomaly detection 
method called kernel outlier detection (KOD), designed to 
address the challenges of outlier detection in 
high-dimensional settings. 
Our research was motivated by limitations of existing methods, 
some of which assume underlying distribution types, or rely 
on hyperparameters that are hard to tune, or struggle when 
data exhibit nonlinear structures, or whose computation 
does not scale well for larger datasets. 
By combining kernel transformations with a new ensemble of 
projection directions, KOD provides 
a flexible and lightweight framework for outlier detection.

From a methodological viewpoint, our work contributes two 
main enhancements. First, we introduced the Basis Vector type
of direction, that is handy in kernel-induced feature spaces. 
And second, we introduced a new way to combine the information 
from different direction types.

Our empirical evaluations on both synthetic and real 
datasets illustrated the effectiveness of KOD. On three 
small datasets with challenging structures we saw how 
different projection direction types capture various 
aspects of the data, with their combination improving 
outlier detection under diverse geometric structures. 
We also studied outlier detection on four large 
benchmark datasets. On the high-dimensional image 
datasets MNIST, MNIST-C, and Fashion MNIST we saw that 
KOD performed among the best methods, illustrating 
its reliability in practical applications.

We end with a cautionary note. In spite of the good
performance of KOD in these examples, there are no
guarantees. Indeed, no method can produce good
results in all situations. This is because the clean 
data can have many different distributions, and the 
outliers can take many forms. What constitutes an
outlier also depends on the kind of analysis
that is being carried out. For instance, a case
may be an outlier in a linear fit but not in a
quadratic fit or a regression tree. In that sense
anomaly detection methods are only a first step,
that should be followed by interpreting its
results.

\vspace{3mm}
\textbf{Acknowledgment.} The comments of two 
reviewers have improved the presentation.

\vspace{3mm}
\textbf{Supplementary Material.} The \proglang{R} 
code of the proposed method and an example script 
are in 
\url{https://wis.kuleuven.be/statdatascience/code/kod_r_code_script.zip}\,.

\vspace{3mm}
\begin{center}
\Large{\textbf{Appendix}}\\
\end{center}
\vspace{1mm}

\noindent {\large{\bf A. An additional dataset}}

\begin{figure}[H]
\centering
\vspace{2mm}
\includegraphics[width=0.40\linewidth]
  {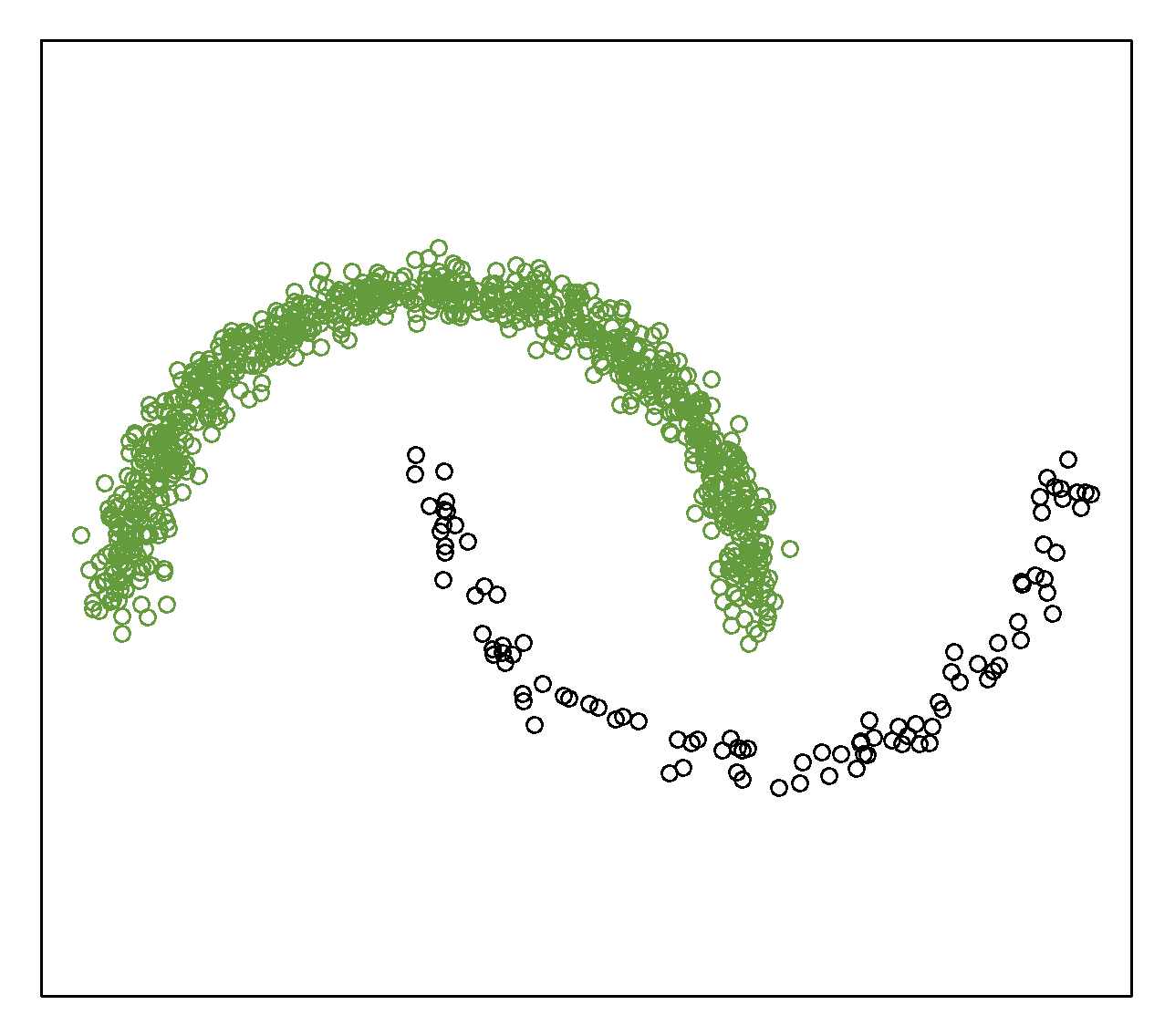}\\
\vspace{-2mm}
\caption{The Moons dataset.}
\label{fig:moons_data}
\end{figure}

\vspace{2mm}
In addition to the `toy' datasets
\textit{Salt-Pepper Ring}, \textit{Circle-Cluster}
and \textit{Inside-Outside} we also considered
another, the \textit{Moons} data. This dataset 
consists of two interleaving half moon shapes,
as seen in Figure~\ref{fig:moons_data}. 
It is widely used to 
demonstrate the limitations of linear separation
and to motivate the use of kernel-based approaches. 
Here the upper half circle (in green) contains most 
of the cases, while the lower one (in black) has 
10\% of the cases. The latter are therefore 
outlying relative to the majority.

\begin{figure}[H]
\centering
\includegraphics[width=1\linewidth]
  {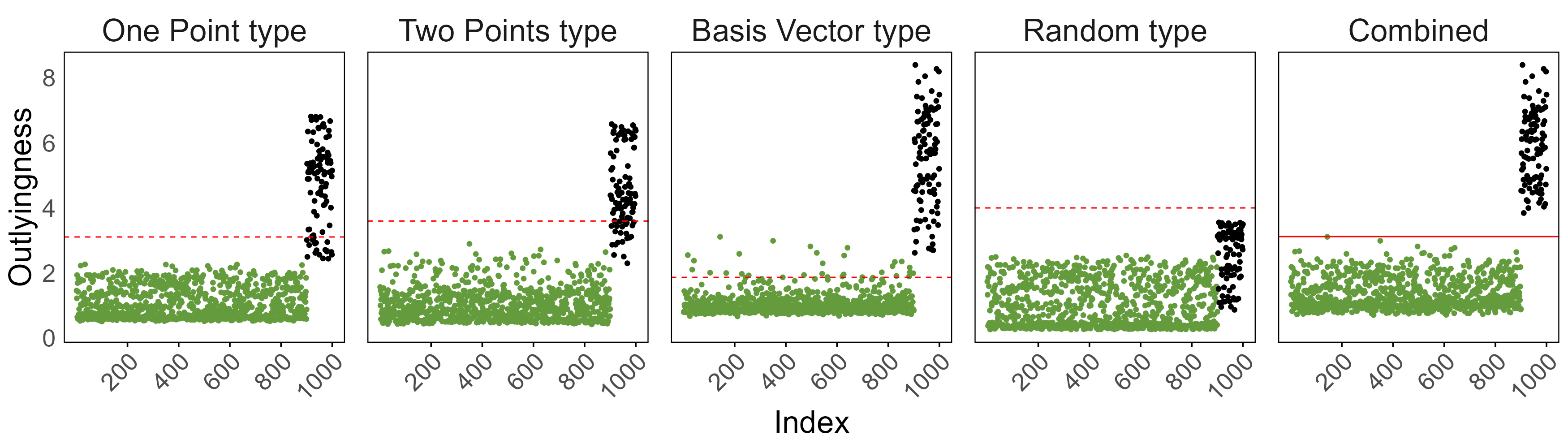}\\
\vspace{-2mm}
\caption{Outlyingness values obtained from each 
direction type and the combined KO outlyingness on 
the \textit{Moons} dataset with 10\% contamination.}
\label{fig:moons_outl1}
\end{figure}

Let us now apply the KOD method, starting with
the RBF kernel. Figure~\ref{fig:moons_outl1} shows 
some results. The first four panels display the 
normalized outlyingness values\linebreak 
$\text{outl}_\text{type}/
\med_j(\text{outl}_\text{type}(\bhfj))$ for each 
type of direction. The last panel shows their 
maximum, which is the final KO. 

In Figure~\ref{fig:moons_outl1} we see that the
Random directions failed to separate the outliers
from the regular points. The other three types
did better, but none of them in a perfect way.
The combined KO values in the last panel do 
separate the outliers from the inliers, with 
the cutoff line nicely in between. 

\begin{figure}[H]
\centering
\includegraphics[width=1\linewidth]
  {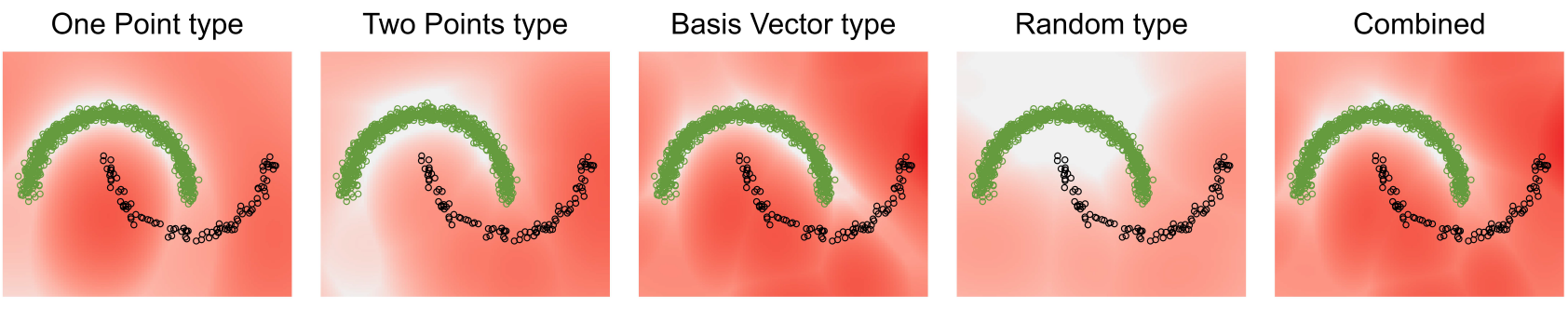}\\
\vspace{-1mm}
\caption{Heatmaps of the outlyingness values 
obtained from each direction type and the 
combined KO, on the \textit{Moons} dataset 
with 10\% contamination.}
\label{fig:moons_grid1}
\end{figure}

The differences between the direction types can 
also be seen in the heatmaps of the SDO in
Figure~\ref{fig:moons_grid1}. The white 
region in each shows where points would be
considered regular, and it varies greatly
across direction types. For instance,
Random considers part of the lower half circle 
as regular. Fortunately the heatmap of the
combination, in the rightmost panel, is
quite accurate.\\

\noindent {\large{\bf B. Additional table with
           Matthews correlation}}

\vspace{3mm}
In the main text the comparisons between methods 
were reported using the P@N measure given by (14). 
Here is an additional table in which the 
performance is instead measured by the Matthews 
correlation of~(13). But now we can only make
comparisons between KOD and the three other methods 
that provide cutoff values. 

\begin{table}[!ht]
\caption{Comparison of outlier detection methods 
applied to four real datasets, with varying 
contamination percentage. The entries are the 
averaged MCC.}
\label{tab:methods_comparison_image_MCC}
\begin{center}
\renewcommand{\arraystretch}{0.8} 
\begin{adjustbox}{width=\textwidth,center}
\begin{tabular}{lrrrrrrrrrrrr}
\toprule
  & \multicolumn{3}{c}{\textbf{MNIST}} & \multicolumn{3}{c}{\textbf{MNIST-C}} & \multicolumn{3}{c}{\textbf{Fashion MNIST}} 
  & \multicolumn{3}{c}{\textbf{PageBlocks}} \\
  \cmidrule(lr){2-4} \cmidrule(lr){5-7} \cmidrule(lr){8-10} \cmidrule(lr){11-13} 
  & \multicolumn{3}{c}{Contamination} & \multicolumn{3}{c}{Contamination} & \multicolumn{3}{c}{Contamination}  & \multicolumn{3}{c}{Contamination}  \\
  & 5\% & 10\% & 20\% & 5\% & 10\% & 20\% & 5\% & 10\% & 20\% & 5\% & 10\% & 20\% \\
\midrule
        
\textbf{KOD} & 0.44 & 0.45 & 0.33 & 0.46 & 0.47 & 0.41 & 0.42 & 0.44 & 0.41 & 0.26 & 0.34 & 0.37 \\
        
\textbf{KRPD} & 0.31 & 0.33 & 0.20 & 0.36 & 0.35 & 0.24 & 0.43 & 0.41 & 0.21 & 0.27 & 0.36 & 0.44 \\

\textbf{KMRCD} & 0.31 & 0.34 & 0.28 & 0.35 & 0.35 & 0.34 & 0.35 & 0.37 & 0.30 & 0.20 & 0.33 & 0.27 \\
        
\textbf{OCSVM} & 0.21 & 0.28 & 0.34 & 0.10 & 0.14 & 0.19 & 0.20 & 0.28 & 0.38 & -0.02 & -0.04 & -0.07 \\     
\bottomrule
\end{tabular}
\end{adjustbox}
\end{center}
\vspace{-5mm}
\end{table}

Table~\ref{tab:methods_comparison_image_MCC} 
shows the MCC results for the four real datasets.
The first thing we notice is that the MCC values
are quite low overall, because here we did not
standardize each column by its maximal value.
The datasets are indeed quite challenging due
to their complexity, and as mentioned before 
there is no `absolute ground truth' for real data. 
The performance of KOD relative to the other
three methods with a built-in decision rule for 
flagging outliers is qualitatively similar to 
the situation for the P@N measure. We see that
OCSVM struggled a bit with MNIST-C and 
PageBlocks.\\

\noindent {\large{\bf C. Computation time}}

\vspace{3mm}
We could not really compare the computation 
times of the methods in the study because 
some were coded in Python, some in 
\textsf{R}, and some used \textsf{R} with 
certain components in 
compiled C++. Instead we investigated the 
computation time of the proposed method on
the three image datasets. These runs were 
part of the simulation leading to Table~2
in the main text and 
Table~\ref{tab:methods_comparison_image_MCC}
here. The data dimension was the number of
pixels per image, yielding $p = 196$, 
$p=784$ and $p=3136$. We varied the
sample size over $n = 250, 500, 1000, 2000$.

\begin{figure}[ht]
\centering
\includegraphics[width=0.8\linewidth]
  {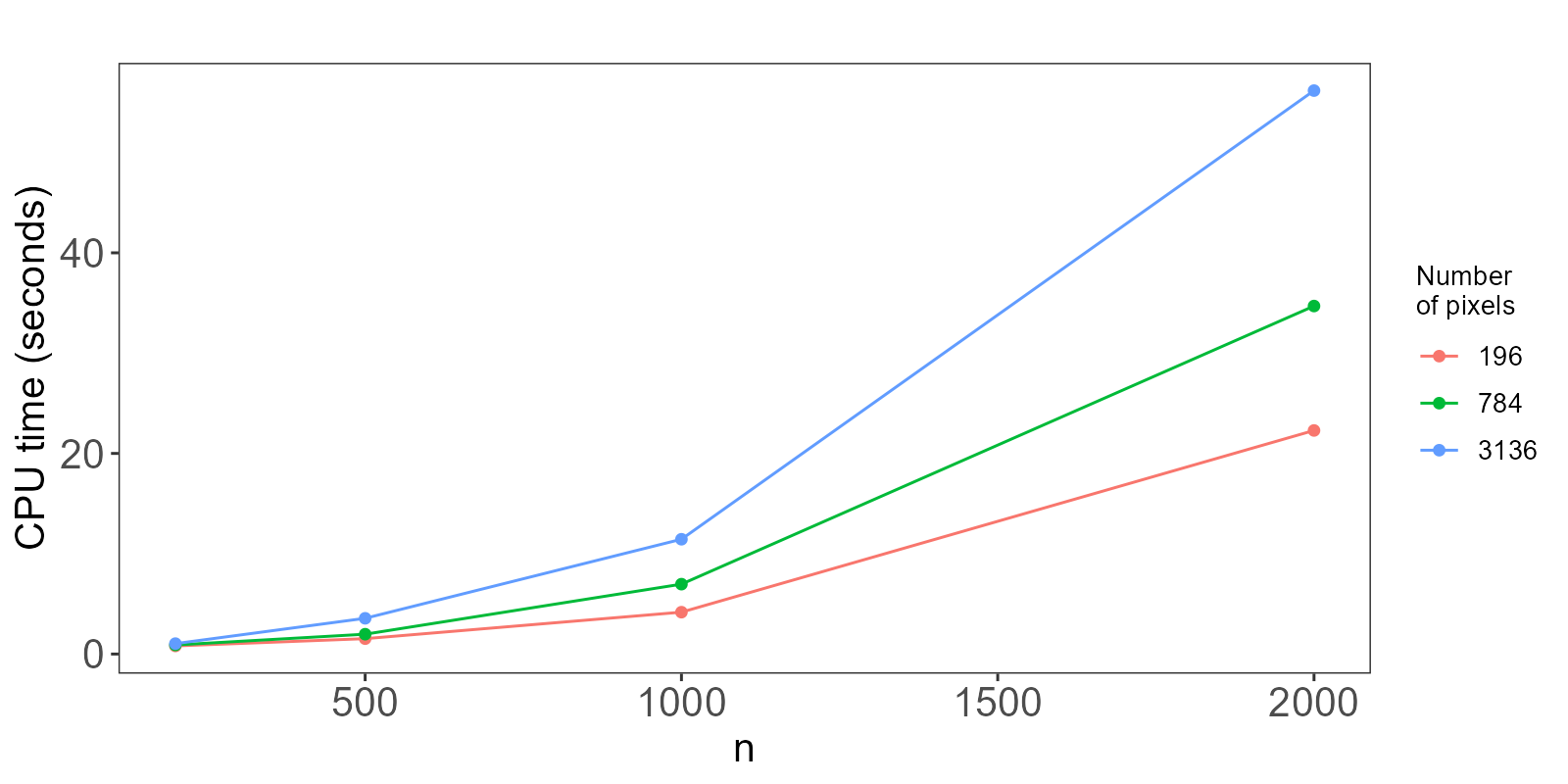}
\caption{Computation time of KOD. The wall-clock 
time is shown in function of the sample size $n$. 
The three curves correspond to different 
dimensions. Experiments were run under Windows 11 
Enterprise on an HP EliteBook 865 G10 notebook 
equipped with an AMD Ryzen 7 PRO 7840U CPU 
(8 cores and 16 threads, 3.3 GHz base) and 
32 GB RAM.}
\label{fig:cpu_time}
\end{figure}

Figure~\ref{fig:cpu_time} shows the wall-clock
time in function of the sample size, with
separate curves for the dimensions $p = 196$ 
(red), $784$ (green) and $3136$ (blue).
As expected we see that the computation time 
goes up with $n$ and $p$, but it remains 
feasible. Speedups are possible by rewriting
parts of the \textsf{R} code in C++.\\

\end{document}